\documentclass[letterpaper, 10 pt, conference]{ieeeconf} 

\IEEEoverridecommandlockouts
\usepackage{tabularx, graphicx,
  amsmath,amsfonts,siunitx,pifont,amssymb}
\usepackage[table]{xcolor}
\usepackage{xcolor, bm, float, multirow, multicol, courier, mathtools, url}
\usepackage[ruled,vlined]{algorithm2e} \pdfminorversion=4
\let\labelindent\relax \usepackage{enumitem} \usepackage{graphicx}
\usepackage{adjustbox}
\usepackage{subcaption}
\usepackage{todonotes}

\newcommand{\etal}{\textit{et al}.~}
\newcommand{\ie}{\textit{i}.\textit{e}.}

\usepackage{pdfpages}
\usepackage{cite}
\usepackage{balance}
\usepackage{hyperref}

\def\secref#1{Sec.~\ref{#1}}
\def\figref#1{Fig.~\ref{#1}}

\def\eqref#1{Eq.~(\ref{#1})}

\title{\LARGE \bf{Elastic and Efficient LiDAR Reconstruction \\ for Large-Scale
Exploration Tasks}}

\author{Yiduo Wang\textsuperscript{1}, Nils Funk\textsuperscript{2}, Milad
Ramezani\textsuperscript{1}, Sotiris Papatheodorou\textsuperscript{2}, Marija
Popovi\'{c}\textsuperscript{2}, Marco Camurri\textsuperscript{1}, \\Stefan
Leutenegger\textsuperscript{2} and Maurice Fallon\textsuperscript{1}
	\thanks{This research is supported by the ESPRC ORCA Robotics Hub (EP/R026173/1). M. Fallon is supported by a Royal Society University Research Fellowship and S. Papatheodorou by the President's PhD Scholarship. N. Funk's PhD is funded by SLAMcore Ltd.
	}
	\thanks{\textsuperscript{1} These authors are with the Oxford Robotics Institute, University of Oxford, UK.
		{\tt\small \{ywang, milad, mcamurri, mfallon\}@robots.ox.ac.uk}}%
	\thanks{\textsuperscript{2} These authors are with the Smart Robotics Lab, Department of Computing, Imperial College London, UK.
    {\tt\small \{nils.funk13,  s.papatheodorou18, mpopovi1,
s.leutenegger\}@ic.ac.uk}}%
}


\begin{document}
	
\setlength{\abovedisplayskip}{4pt}
\setlength{\belowdisplayskip}{4pt}
	
\maketitle 
\thispagestyle{empty} 
\pagestyle{empty}
	

\begin{abstract}
We present an efficient, elastic 3D LiDAR reconstruction
framework which can reconstruct up to maximum LiDAR ranges (\SI{60}{\meter}) at
multiple
frames per second, thus enabling robot exploration in large-scale
environments. Our approach only requires a CPU. We focus on three
main challenges of
large-scale reconstruction: integration of long-range LiDAR
scans at high frequency, the
capacity to deform the reconstruction after loop closures are detected, and 
scalability for long-duration
exploration. Our system extends upon a state-of-the-art
efficient RGB-D volumetric
reconstruction technique, called supereight, to
support LiDAR scans and a newly developed submapping technique to allow
for dynamic correction of the 3D reconstruction. 
We then introduce a novel pose graph clustering and submap fusion
feature to make the proposed system more scalable for large environments. 
We evaluate the performance using two public datasets including outdoor exploration
with a handheld device and a drone, 
and with a mobile robot exploring an underground
room network. 
Experimental results demonstrate that our system can reconstruct at 
\SI{3}{\hertz} with \SI{60}{\meter} sensor range and \SI{\sim5}{\centi\meter}
resolution, while state-of-the-art approaches can only reconstruct to
\SI{25}{\centi\meter} resolution or \SI{20}{\meter} range at the same
frequency. 
\end{abstract}
	
\section{Introduction}

Dense reconstruction is an active research topic.
Being able to recover rich geometric information in real time is important for 
applications such as active mapping~\cite{Bircher2018, Dai_ICRA2020}, obstacle 
avoidance~\cite{oleynikova2017voxblox} and industrial inspection~\cite{Hollinger2013, Franz2016}. 
Lower cost and denser LiDAR sensors have come to the market thanks to the focus on 
self-driving car research. However, large scale exploration and reconstruction
still remain challenging problems. 

\begin{figure}[t!]
    \centering
    \includegraphics[width=0.9\columnwidth]{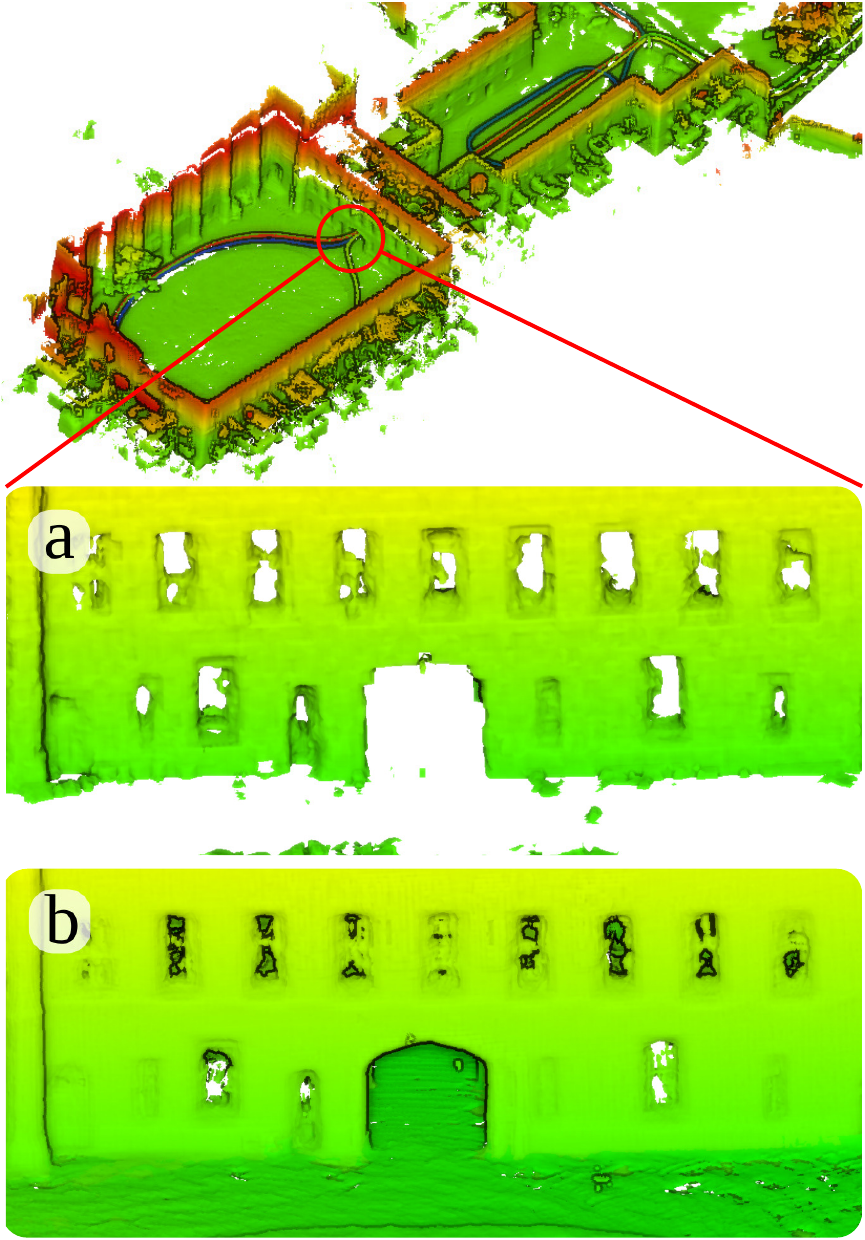}
    \caption{\small{Exploration trajectory and 3D reconstruction result from our elastic
\textit{supereight}
multi-resolution TSDF pipeline on Newer College Dataset. The close-ups focus on the narrow tunnel on
the opposite side of the Quad area from experiment's start. (a - the first
submap \SI{40}{\meter}
away; b - revisiting after a large loop closure.) More demonstrations are available in the supplementary video: \url{https://youtu.be/13xz4YnCGW0}}}
    \label{fig:ncd_short_tunnel}
    \vspace{-7mm}
\end{figure}

A major challenge in reconstruction is global consistency
because the accumulation of some degree of odometry error 
is unavoidable during large-scale exploration ---
even for high performance approaches such as LOAM~\cite{zhang2014loam}. 
Error increases with distance travelled and is
typically corrected by loop closures in the context of SLAM.
These corrections can also be applied to the map representation used within SLAM systems,
which are typically point clouds for LiDAR-based SLAM. 
However, point cloud representations lack the occupancy information
necessary for exploration and planning to distinguish between unknown and observed-free voxels.
If volumetric or surface mesh reconstructions are built on-the-fly
by an exploring robot,
the resulting reconstruction would be rigid and impossible to incorporate
the effect of loop closures.


Another challenge is finding a good trade-off between the resolution/scale
of the reconstruction, and the speed/efficiency of the system.
A precise representation of occupancy is important for path planning, especially
when navigating through tunnels and door ways such as in \figref{fig:ncd_short_tunnel}.
In this work, we greatly expand \textit{supereight},
a state-of-the-art 
multi-resolution reconstruction framework for RGB-D cameras, 
to incorporate 3D multi-beam LiDAR scans.
It uses high resolution in close proximity for path 
planning and lower resolution at longer ranges for fast reconstruction. 
We then demonstrate that our approach is more efficient than
other state-of-the-art pipelines at high resolutions.

Finally, we also implement a novel system to cluster the SLAM pose graph and reconstruction,
inspired by large-scale systems such as the \textit{Atlas} SLAM
framework~\cite{Bosse2003} and SLAM pose graph sparsification techniques \cite{Carlevaris2013,
Mazuran2014NonlinearGS, Vallve2018}.
In our work, we focus on improving the scalability of mapping instead of pose-graph SLAM.
Reconstructions of the same physical space are fused together to avoid redundant mapping. 

The contributions and features of our research are the following:
\begin{itemize}
    \item An elastic 3D reconstruction system that supports corrections to its underlying shape, e.g. from loop closures. 
    \item A pose graph clustering and submap fusion strategy that makes the
reconstruction's memory usage grow proportionally with the size of
the environment rather than the duration of exploration.
    \item Incorporation of LiDAR into a state-of-the-art reconstruction framework 
for occupancy mapping, which achieves
multi-fps (\SI{3}{\hertz}) full range (\SI{60}{\meter}) LiDAR scan integration with high  resolution
(\SI{\sim5}{\centi\meter}) to enable high precision motion planning and long
range
autonomy.  To the best of our knowledge, this is the first system that achieves
this level of performance.
    \item Evaluation of the system using real-world datasets in large-scale environments against
state-of-the-art methods such as Octomap and Voxgraph.
\end{itemize}

The remainder of this paper is organised as follows. In Section
\ref{sec:RelatedWork} we discuss the
related work. Section \ref{sec:Supereight} focuses on the reconstruction method and Section 
\ref{sec:Elasticity} explains how we achieve elasticity. Experimental results are presented in Section 
\ref{sec:Experiments}. Section \ref{sec:FutureWork} discusses conclusions
and future work.

\section{Related Work}
\label{sec:RelatedWork}

Dense SLAM and mapping systems are very active areas of research. 
Our proposed system mainly focuses on two aspects, namely large-scale 
dense reconstruction, and submaps and elasticity. In this section, we 
will give a brief review of the most relevant systems. 

\paragraph{Dense Reconstruction}
There is a variety of representations for 3D environments. 
KinectFusion~\cite{newcombe2011kinectfusion} and
ElasticFusion~\cite{whelan2015elasticfusion} used
Truncated Signed Distance Function (TSDF) and surfel respectively
to achieve highly detailed reconstructions,
utilising GPU to integrate inputs from RGB-D cameras. 
These sensors typically have a myopic sensing range of only \SI{3}{\meter}
as compared to as much as \SI{100}{\meter} for terrestrial LiDAR.
Recent work by Park~\etal\cite{park2020} revised the dense surfel model 
for dense LiDAR SLAM. 

Another technique to improve reconstruction scalability is dynamic
allocation via data structures such as octrees~\cite{Ho2018, Sodhi2019, duberg2020ufomap}
and hash-tables~\cite{Klingensmith2015, Dai2017}.
For instance, Niessner~\etal\cite{Niessner2013} employed Hashing Voxel Grid (HVG)
to exploit environment sparsity for large-scale RGB-D mapping using TSDF, 
which was also incorporated in 
systems such as BOR$^2$G~\cite{Tanner2018} and InfiniTAM~\cite{Kahler2015, Kahler2016}.

Systems designed for navigation and motion planning prefer to use
occupancy grids for dense reconstruction to explicitly represent 
free and unknown space~\cite{DeGregorio2017, Ho2018, Sodhi2019, duberg2020ufomap}. 
Vespa~\etal\cite{Vespa2019Supereight} presented \textit{supereight}, a
volumetric SLAM system for RGB-D measurements 
that uses an efficient octree structure to store 
either TSDF or Occupancy information 
at various levels of detail depending on the distance to surfaces. 
Compared to InfiniTAM~\cite{Kahler2015}, \textit{supereight}
demonstrated better efficiency.
In this work, we extend \textit{supereight} to support the much
longer ranges from 3D LiDAR sensors, while also utilising
the adaptive resolution feature.

\paragraph{Submaps and Elasticity}
Odometry error accumulated through incremental tracking methods 
can lead to distortion in the map.
Dense SLAM systems such as KineticFusion~\cite{newcombe2011kinectfusion} and 
ElasticFusion~\cite{whelan2015elasticfusion}
tightly couple their reconstructions with the SLAM trajectory
and correct individual past scans to bend the mapped reconstruction upon loop closure.  
This method, however, suffers from poor scalability in large-scale online operations. 

Another technique to improve global consistency is to represent the full
reconstruction as a collection of submaps with bounded size.
This technique originated in SLAM research such as the \textit{Atlas} framework 
by Bosse~\etal\cite{Bosse2003} and DenseSLAM by Nieto~\etal\cite{Nieto2006DenseSLAM}. 
Ho~\etal\cite{Ho2018} and Reijgwart~\etal\cite{reijgwart2020voxgraph} both 
exploited submaps to achieve elasticity in dense reconstruction. 
Their reconstructions are based on OctoMap~\cite{Hornung2013OctoMap} 
and Voxblox~\cite{oleynikova2017voxblox}, respectively.
Upon loop closure, submaps in both systems can be moved around to keep global consistency.


Our approach is most directly motivated by the work of 
Reijgwart~\etal\cite{reijgwart2020voxgraph}. 
Voxgraph incorporates both LiDAR and RGB-D measurements, and was demonstrated to be
lightweight enough to run on a drone.
The main drawback of this system is 
the dense data structure of hash-tables. TSDF integration and ESDF computation 
are all carried out at the finest resolution. 
The typical parameter setup the authors used for a LiDAR exploration task was
short range (\SI{16}{\meter})
with coarse resolution (\SI{20}{\centi\meter}). With such parameters, it would
be impossible to plan at full LiDAR ranges.


\section{Large-scale Supereight}
\label{sec:Supereight}

\begin{figure*}[!h]
    \centering
    \includegraphics[width=.9\linewidth,trim={0cm 0cm 0cm
        0cm},clip]{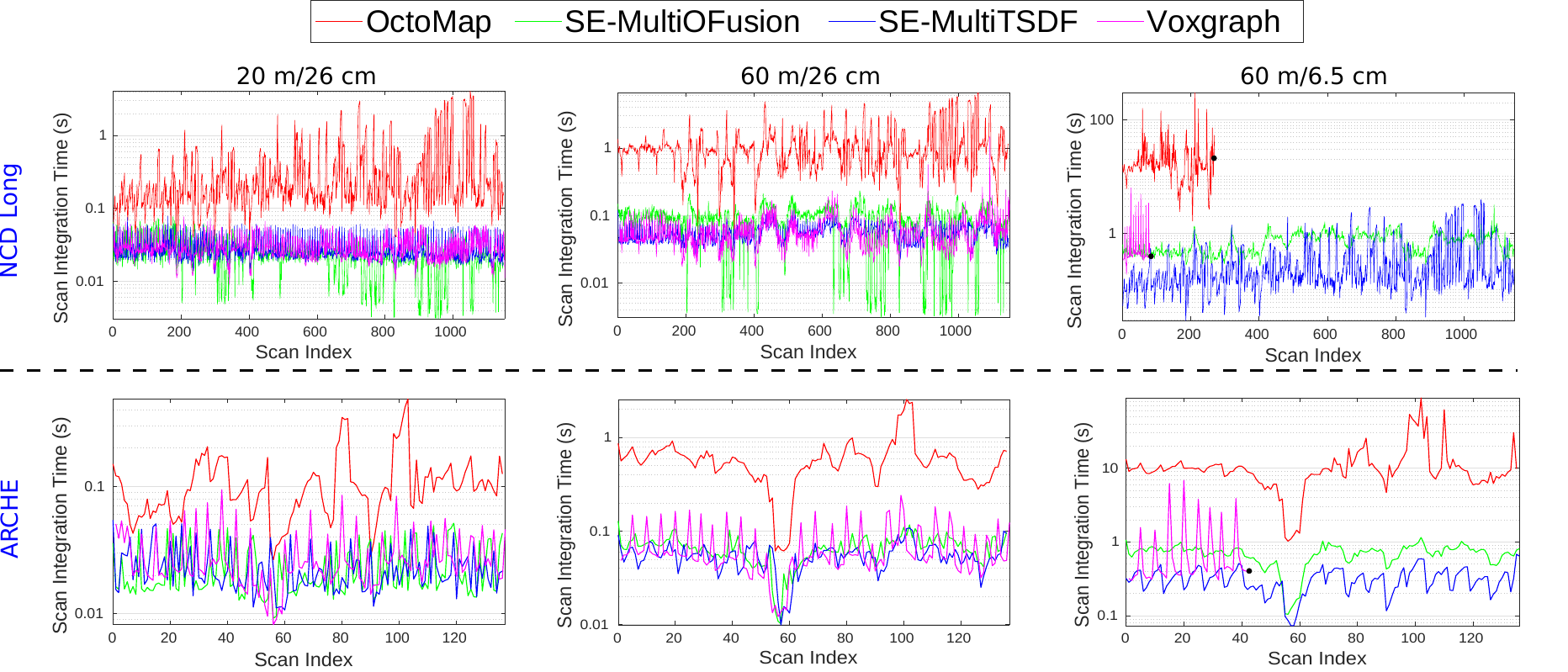}\
    \caption{\small{Integration time per LiDAR scan of different reconstruction systems in
            large-scale exploration experiments. Our goal is to achieve high resolution at maximum sensor range
            (right column).}}
    \label{fig:integrationtime}
    \vspace{-5mm}
\end{figure*}

In this section, we describe the core reconstruction component
of our system, \textit{supereight}~\cite{Vespa2019Supereight},
and the significant improvements needed to support long
range LiDAR sensing.~\textit{Supereight} is a volumetric,
octree-based SLAM 
pipeline with adaptive resolution that uses Morton codes to achieve efficient spatial
octree traversals. Instead of individual voxels, \textit{supereight} stores blocks which
aggregate $8 \times 8 \times 8$ voxels as the finest leaves of the tree.
This results in fewer memory allocations and improved cache 
locality during updates, improving performance.
It is also capable of integrating data at different octree
levels, further increasing efficiency.

In this work, 
the mapping component of \textit{supereight} is connected 
to a LiDAR SLAM system~\cite{Ramezani2020LiDARSLAM}. 
We expand both 
\textit{supereight}'s multi-resolution TSDF 
(\textit{MultiresTSDF})~\cite{Vespa2019Supereight} and 
multi-resolution occupancy (\textit{MultiresOFusion})~\cite{funk2020multiresolution} 
pipelines to incorporate LiDAR inputs.
The original RGB-D \textit{supereight} uses a pinhole camera model. 
To incorporate LiDAR data into the framework, 
organised LiDAR point clouds are converted to depth images,
and the projection model is approximated with a spherical camera model
by defining a pair of azimuth and elevation angles 
for each pixel in the depth image
based on sensor specification.
Compared to the pinhole camera model, 
the LiDAR model incorporates a longer range and larger Field of View (FoV), but the distance
measurements are sparser.
We also create local submaps in this system
to replace the single global map in the original pipeline. 
This is further explained in \secref{sec:Elasticity}. 

\subsection{Multi-resolution}
\label{subsec:DataStructure}

Long range measurements from a LiDAR
cover a much larger amount of free space than an RGB-D camera.
\textit{Supereight} can update the octree at various levels 
depending on the effective resolution of the sensor,
by updating cubes consisting of several voxels 
instead of individual voxels.
The benefit of this approach is a reduction in the number of 
octree updates, 
resulting in reduced integration time~\cite{Vespa2019Supereight}.
This performance increase is especially important in the case of 
LiDAR sensors where a single scan may contain measurements ranging 
from a few meters to \SI{60}{\meter} away. 

Due to a larger FoV, it is more likely for LiDAR rays to hit surfaces at shallow angles
than those of RGB-D cameras, which results in aliasing artefacts.
In this work, we update \textit{supereight}'s integration level selection
method for a particular depth measurement to make it suitable for LiDAR sensors. 
This update reduces aliasing artefacts,
increases speed and decreases memory consumption at large distances.
For each ray  $\mathbf{r}$,
we consider the minimum angle between 
two adjacent LiDAR scan rays, which in turn defines a circular cone. 
We then determine the scale of update volume at a distance measurement $d_r$ 
the largest block of voxels that fits inside this cone, 
up to $8 \times 8 \times 8$ voxels. 
Thus, measurements can be integrated into volumes at adaptively selected
resolutions. 

We use the propagation strategies described 
in~\cite{Vespa2019Supereight} and~\cite{funk2020multiresolution}
for MultiresTSDF and MultiresOFusion, respectively,
to keep the hierarchy consistent between different 
integration levels. 
Additionally, in MultiresOFusion
the maximum occupancy and observed state at the 
finest integration level are up-propagated to each parent 
level
to provide fast occupancy queries at different levels.
MultiresOFusion also explicitly keeps track of
free space at the coarsest possible scale while 
preserving details about unknown space.

\subsection{LiDAR Integration}
\label{subsec:LiDARIntegration}

\textit{LiDAR integration} refers to the process of creating 
a new reconstruction or updating an existing one based on a new scan.
The LiDAR used in our experiments is an Ouster OS1-64. 
It produces organised dense point clouds
of $64\times1024$ points at \SI{10}{\hertz} (\ie ~655k points/s), with a
vertical FoV of \SI{33.2}{\degree} and a horizontal FoV of
\SI{360}{\degree}. 
Scans are converted from point clouds to spherical range images to facilitate their 
inclusion into \textit{supereight}.

The MultiresOFusion pipeline stores the occupancy probability
in log-odds form which results in free, unknown and occupied voxels 
having negative, zero and positive log-odds values, respectively.
Occupancy update follows the convention of 
adding a new log-odds measurement~\cite{Hornung2013OctoMap,Vespa_RAL2018}.
The log-odds measurement along a ray
is a distance-dependent piecewise linear function
explained in detail in~\cite{funk2020multiresolution}. 


The integration process for the MultiresTSDF pipeline
is described in~\cite{Vespa2019Supereight}.
To avoid artefacts when using long-range LiDAR scans,
we modify the pipeline so that the TSDF truncation bound adapts to 
the integration level.

\subsection{Runtime Performance}
\label{subsec:Runtime}

\begin{figure}[t]
    \centering
    \includegraphics[width=\linewidth,trim={0cm 0cm 0cm 0cm},clip]{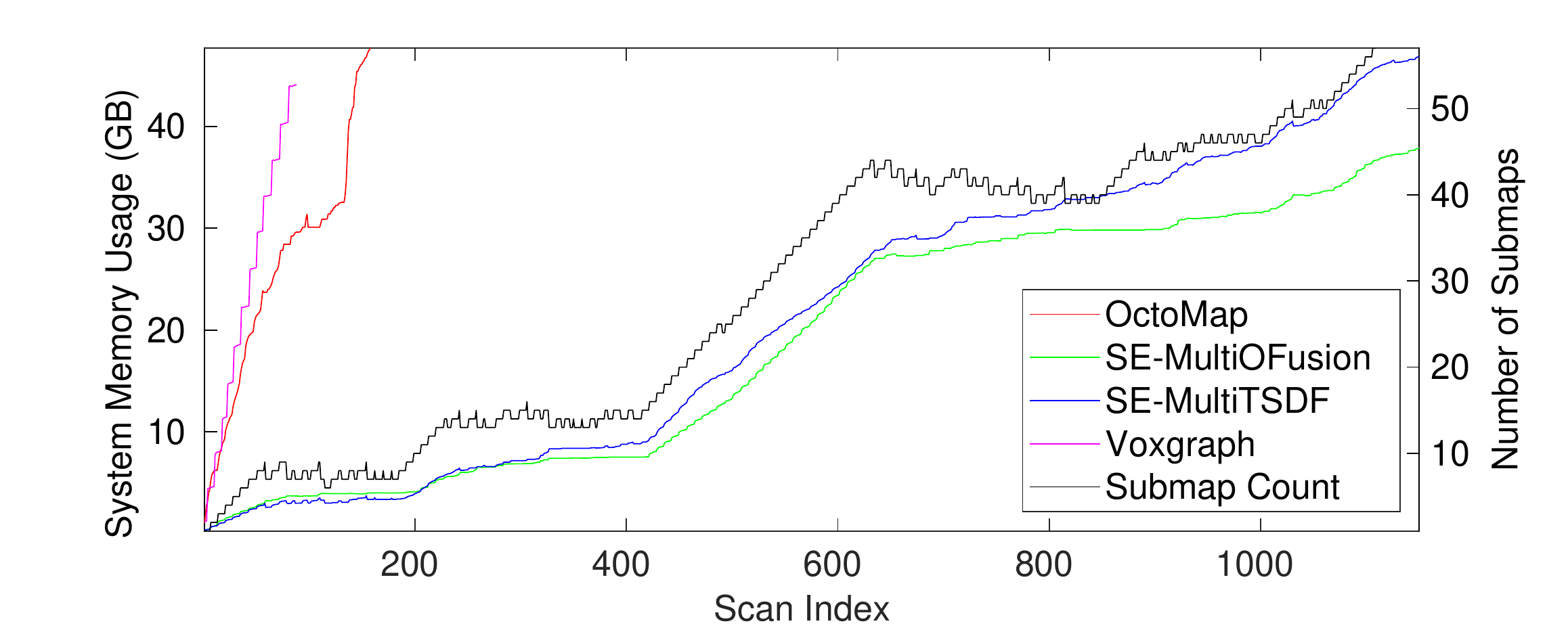}
    \includegraphics[width=\linewidth,trim={0cm 0cm 0cm 0cm},clip]{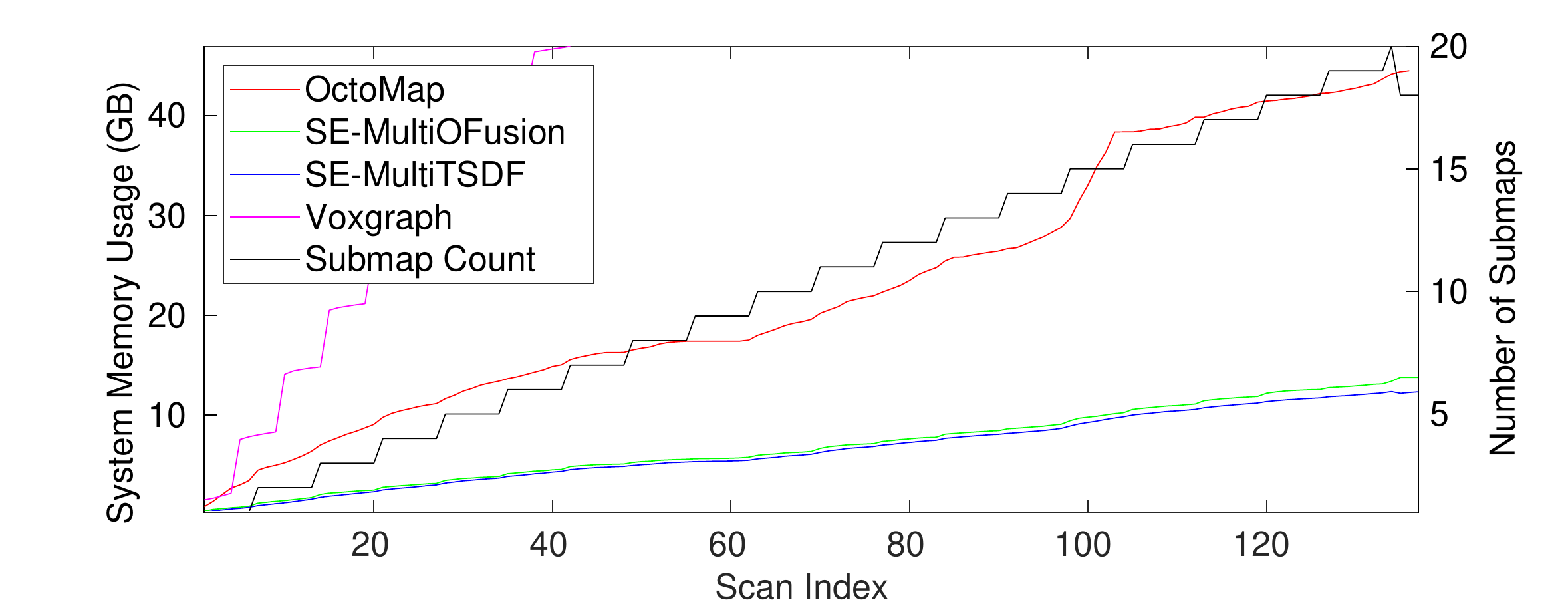}
    \caption{\small{Memory usage of each pipeline in the NCD Long (top) and ARCHE (bottom)
            experiments with
            \SI{60}{\meter} range and \SI{6.5}{\centi\meter} resolution. Memory usage of our pipelines had a
            non-linear profile in NCD Long because of the submap fusion feature 
            (\secref{subsec:Fusion}).}}
    \label{fig:memory_usage}
    \vspace{-5mm}
\end{figure}

To evaluate our system's efficiency when integrating LiDAR scans, we tested it using 
the Newer College Dataset (NCD)~\cite{ramezani2020newer} 
and the dataset made available with~\cite{reijgwart2020voxgraph} (ARCHE).
NCD consists of two experiments with different durations, 
and we present in this paper the \SI{44}{\minute} one (NCD Long).
Both datasets were of large-scale
(approximately \SI{135x225}{\meter\squared} for NCD and \SI{70x160}{\meter\squared} for ARCHE) 
with an Ouster OS1-64 LiDAR and a RealSense camera (D435i for NCD and D415 for ARCHE).

The baseline algorithms we chose for comparison are OctoMap~\cite{Hornung2013OctoMap} and
Voxgraph~\cite{reijgwart2020voxgraph}, to assess the respective Occupancy and TSDF pipelines. 
For these experiments, we fed one point cloud every \SI{2}{\meter}
travelled into the
reconstruction systems.
All reconstruction computations were performed on a laptop with an
Intel\textsuperscript{{\textregistered}}
Xeon E3-1505M v6 CPU, 16 GB of RAM and 32 GB of swap memory.

We evaluated the computation time using three different sets of 
maximum scan range and voxel resolutions:
\begin{itemize}
\item \SI{20}{\meter} max range with \SI{26}{\centi\meter} resolution
\item \SI{60}{\meter} max range with \SI{26}{\centi\meter} resolution
\item \SI{60}{\meter} max range with \SI{6.5}{\centi\meter} resolution
\end{itemize}

\figref{fig:integrationtime} shows the integration time at the different
range/resolution combinations for NCD Long and ARCHE experiments 
(top and bottom rows, respectively).
We focus on mapping at high resolution (\SI{6.5}{\centi\meter}) with 
maximum LiDAR range (\SI{60}{\meter}), which is presented in the right 
column of \figref{fig:integrationtime}. 
In both experiments, Voxgraph terminated early due to memory limits, 
as did OctoMap in NCD Long.
\figref{fig:memory_usage} shows the memory consumption of each pipeline, 
as well as the growth of submaps in the proposed system.
The memory usage of OctoMap and Voxgraph increases more quickly than
both \textit{supereight} pipelines, 
thus illustrating how \textit{supereight}'s multi-resolution improves the 
memory efficiency of the reconstruction, 
allowing it to scale to larger environments.

Overall, OctoMap is the least efficient of the evaluated methods. With coarse
resolutions, Voxgraph exhibits similar performance to \textit{supereight}.
However, at \SI{6.5}{\centi\meter} resolution, the
MultiresTSDF pipeline
is faster than Voxgraph,
while MultiresOFusion is on a par with Voxgraph.


\begin{figure}[t]
    \centering
    \includegraphics[width=\linewidth]{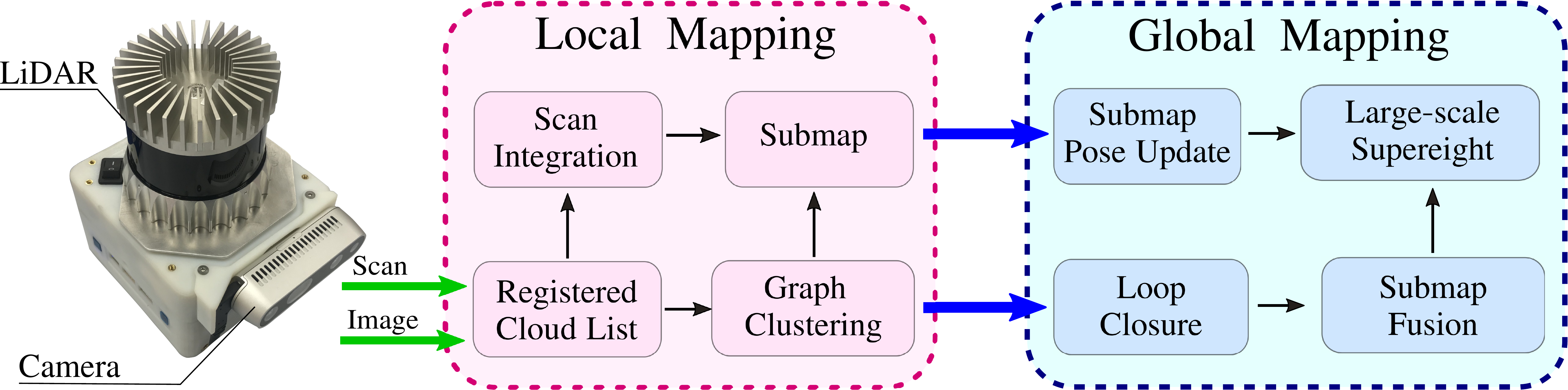}
    \caption{\small{An overview of the system.
Large-scale Supereight is the dense reconstruction core described in \secref{sec:Supereight}, 
and all other components are specially designed for the proposed system.
Registered Cloud List is the input. Local mapping clusters the SLAM pose graph
and integrates individual LiDAR scans into submaps. 
Global mapping updates poses of submaps upon loop closures 
and fuses overlapping submaps together. See \secref{sec:Elasticity} for more details. }}
    \label{fig:system_overview}
    \vspace{-4mm}
\end{figure}


\section{Submaps and Elasticity}
\label{sec:Elasticity}

In this section, we detail the components of the elastic
reconstruction pipeline, which is shown in \figref{fig:system_overview}.

Our pose graph SLAM system~\cite{Ramezani2020LiDARSLAM} fuses the sensors
signals described in \secref{subsec:Runtime} to compute relative odometry
which is locally consistent with drift rates in the order of 
\SI{1}{\meter} per \SI{100}{\meter} travelled. 
In this way we collect a sequence of point clouds which are registered to one another locally,
as well as a corresponding relative pose estimate for the robot/device.
Upon loop closures, we form a full pose graph.
We call this a \textit{Registered Cloud List}.
The pose graph SLAM system is external to our mapping pipelines, and systems 
such as \cite{zhang2014loam, Kim2018ScanContext} could provide be more advanced options.

In our frame convention, 
the Map frame $\{\mathcal{M}\}$ defines a global fixed frame of reference. The
base frame of the device at time $k$ is defined as $\{\mathcal{B}_k\}$.
The SLAM system provides a pose graph with $Q+1$ nodes $X_k,
k\in\{0,\dots,Q\}$. Each node describes the estimated pose
  of the device  expressed in the map frame
$^\mathcal{M}\mathbf{T}_{\mathcal{B}_k}\in\mathbf{SE}(3)$. The graph's topology
consists of both odometry edges (\ie~connecting two consecutive nodes) and
loop closure edges.

Each node of the graph is associated with a raw point cloud
 from the LiDAR. The point
clouds have a fixed number of points $p\in\mathbb{R}^3$ expressed in the LiDAR
frame $\{\mathcal{L}\}$. Given a node $X_k$ and its associated point cloud
$C_k$, the pose of the LiDAR
$^\mathcal{M}\mathbf{T}_{\mathcal{L}_k}$ can be computed as $
^\mathcal{M}\mathbf{T}_{\mathcal{L}_k} =
{^\mathcal{M}\mathbf{T}_{\mathcal{B}_k}}
{^\mathcal{B}\mathbf{T}_\mathcal{L}}
$, 
where $^\mathcal{B}\mathbf{T}_\mathcal{L}\in\mathbf{SE}(3)$ is a static
transform known by design.

The output of the reconstruction system consists of $N+1$
submaps. Each submap $\mathcal{S}_i, i\in\{0\text{...}N\}$
contains:
\begin{itemize}
    \item The reconstruction as an occupancy map or TSDF
    \item The root pose of $\mathcal{S}_i$
    \item The node indices and scans used to construct $\mathcal{S}_i$
\end{itemize}

A submap's root pose defines the submap's transformation with respect to the map frame $^\mathcal{M}\mathbf{T}_{\mathcal{S}_i}$. 


\begin{figure*}[t]
    \centering
    \includegraphics[width=\linewidth,trim={0cm 0cm 0cm 0cm},clip]{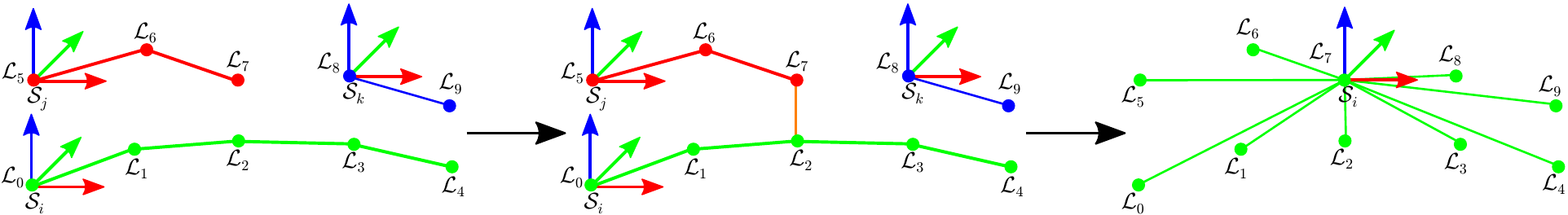}
    \caption{\small{An example of graph clustering 
    and submap fusion based around a loop closure. 
    Nodes $\mathcal{L}_{0:4}$ and $\mathcal{L}_{5:9}$ represent the two traversals of a location
from a pose graph. $\mathcal{L}_{0:4}$ belong to green submap $\mathcal{S}_i$, $\mathcal{L}_{5:7}$
to red submap $\mathcal{S}_j$ and $\mathcal{L}_{8:9}$ to blue submap $\mathcal{S}_k$. Because these
nodes have been grouped together by the graph clustering (\secref{subsec:Sparsification}), these
three submaps are all merged into submap $\mathcal{S}_i$. }}
    \label{fig:sparsification_and_fusion}
    \vspace{-5mm}
\end{figure*}

\subsection{Graph Clustering}
\label{subsec:Sparsification}

The \textit{graph clustering} module processes the pose graph of
the Registered Cloud List and groups graph nodes together into
different submaps.
The clustered graph further guides \textit{scan integration}
(\secref{subsec:ScanIntegration}) and 
\textit{submap fusion} (\secref{subsec:Fusion}). 

To perform clustering, we first divide the pose graph edges into
odometry and loop closure edges. Odometry edges represent constraints between
consecutive pairs of nodes,
while loop closure edges are the constraints between nodes that are distant in
the graph but correspond to similar scans of revisited places.

If there are no loop closures, grouping into submaps is based only on
the odometry chain with a distance $\lambda_\textup{odom}$. In this case, the first node $X_{i,0}$ of
submap $\mathcal{S}_i$ defines the submap's root pose 
$^\mathcal{M}\mathbf{T}_{\mathcal{S}_i}$ with its corresponding LiDAR pose ${^\mathcal{M}\mathbf{T}_{\mathcal{L}_{i,0}}}$.

For the subsequent nodes, we compute the
distance travelled along the pose graph from the root pose. If a new node is
within the distance
threshold $\lambda_\textup{odom}$, the associated LiDAR scan is integrated into
$\mathcal{S}_i$ according to \secref{subsec:ScanIntegration}.
When the new node exceeds the distance threshold, a new submap
$\mathcal{S}_{i+1}$ is spawned with that node. This is based on the assumption that the odometry
drift is proportional to distance travelled.

Upon loop closure, we cluster together nodes that are 
within a threshold $\lambda_\textup{cluster}$ around the pair of nodes that form the closure. 
\figref{fig:sparsification_and_fusion} presents an example of clustering. 
In this example, $\mathcal{L}_2$ and $\mathcal{L}_7$ are 
connected by a loop closure edge. 
We then compute the distances 
from every surrounding node to this loop closure pair
along the pose graph, 
again assuming that odometry drift is proportional to distance travelled. 
In the case of $\mathcal{L}_9$, its distance is computed as: 
\begin{equation}
\begin{split}
d_{\mathcal{L}_7, \mathcal{L}_9} &= d_{\mathcal{L}_7, \mathcal{L}_8} + d_{\mathcal{L}_8, \mathcal{L}_9} \\
&= ||{^\mathcal{M}\mathbf{t}_{\mathcal{L}_7}} - {^\mathcal{M}\mathbf{t}_{\mathcal{L}_8}}||
+ ||{^\mathcal{M}\mathbf{t}_{\mathcal{L}_8}} - {^\mathcal{M}\mathbf{t}_{\mathcal{L}_9}}||
\end{split}\end{equation}
and because $d_{\mathcal{L}_7, \mathcal{L}_9} < \lambda_\textup{cluster}$, 
$\mathcal{L}_9$ is included in this cluster. 
Loop closure clusters guide \textit{submap fusion} in \secref{subsec:Fusion}. 

\subsection{Scan Integration}
\label{subsec:ScanIntegration}

After clustering, a point cloud $C_k$ at node $X_k$ is
integrated into the submap $\mathcal{S}_i$.
We first compute the relative pose between $\mathcal{L}_k$ and $\mathcal{S}_i$:
\begin{equation}
^{\mathcal{S}_i}\mathbf{T}_{\mathcal{L}_k} =
{^\mathcal{M}\mathbf{T}_{\mathcal{S}_i}^{-1}} \;
{^\mathcal{M}\mathbf{T}_{\mathcal{B}_k}} \;
{^\mathcal{B}\mathbf{T}_\mathcal{L}}
 \end{equation}
When $X_k$ is the first node of a submap, its cloud $C_k$ creates the initial
reconstruction of $\mathcal{S}_i$ and $^\mathcal{M}\mathbf{T}_{\mathcal{S}_i} = 
{^\mathcal{M}\mathbf{T}_{\mathcal{L}_k}}$. 

Using $^{\mathcal{S}_i}\mathbf{T}_{\mathcal{L}_k}$, we transform every point $p
\in C_k$ from
the LiDAR frame $\{\mathcal{L}_k\}$ to the submap frame $\{\mathcal{S}_i\}$. 
We then update the submap reconstruction according to~\secref{subsec:LiDARIntegration}.

\subsection{Submap Fusion}
\label{subsec:Fusion}

\textit{Submap fusion} merges the submaps where a loop closure is detected.
For each loop closure
cluster described in \secref{subsec:Sparsification}, we search through all existing submaps and find
those that contain nodes from this cluster. These submaps are
then fused
together as illustrated in \figref{fig:sparsification_and_fusion}.

To fuse the submaps $\mathcal{S}_j$ and $\mathcal{S}_i$, we first need to
transform every voxel of $\mathcal{S}_j$ with coordinates $v_j
\in \mathbb{R}^3$ into the coordinate system of $\mathcal{S}_i$ to obtain $v_i$:
\begin{equation}
\begin{bmatrix}
v_i \\ 1
\end{bmatrix}
= {^{\mathcal{S}_i}\mathbf{T}_{\mathcal{S}_j}}
\begin{bmatrix}
v_j \\ 1
\end{bmatrix}
,\
{^{\mathcal{S}_i}\mathbf{T}_{\mathcal{S}_j}} =
{^\mathcal{M}\mathbf{T}^{-1}_{\mathcal{S}_i}}\;
{^\mathcal{M}\mathbf{T}_{\mathcal{S}_j}}
\end{equation}

If the voxel $v_i$ falls out of the current scanned space in $\mathcal{S}_i$, it
will
be newly allocated and assigned as $v_j$ in $\mathcal{S}_j$.
Otherwise, the voxel data in $v_j$ will be integrated into $v_i$ following the
model in \secref{subsec:LiDARIntegration}.

Submap fusion prevents new submaps from being spawned
when the same space is revisited. Updating an existing submap
is more memory efficient than creating two overlapping submaps.
This has the advantage of making the reconstruction complexity grow
proportionally with the
amount of space explored rather than the duration of the exploration. 

The memory usage of the long range (\SI{60}{\meter}) 
high resolution (\SI{6.5}{\centi\meter}) NCD Long experiment, 
as presented in \figref{fig:memory_usage}, 
demonstrates such benefit. 
\figref{fig:memory_usage} also presents the growth of submaps in the experiment. 
Up to scan 400 in the NCD Long experiment, the number of submaps has limited
growth because the experiment stays within the Quad area and loops are closed.
There after the device explored the wide open Parkland area - with submap growth becoming linear.
Ideally submap growth should have fully plateaued when revisiting the same area regularly (after scan 600). 
Improving submap reduction to achieve the plateauing is a future work. 

\subsection{Submap Pose Update}
\label{subsec:Update}

The \textit{submap pose update} module ensures global
consistency in the reconstruction. When loop closure occurs, the pose graph 
and poses of the SLAM system are updated. 

The naive approach of updating all the submaps upon loop closure is
computationally infeasible for real-time applications, 
as discussed by Sodhi~\etal\cite{Sodhi2019}. 
Instead, we define a criterion to determine whether a submap $\mathcal{S}_i$ needs
to be corrected, such that a large-scale reconstruction can be
selectively and efficiently updated. 
Let $^\mathcal{M}\hat{\mathbf{T}}_{\mathcal{S}_i}$ denote the updated
transformation $^\mathcal{M}\mathbf{T}_{\mathcal{S}_i}$ of $\mathcal{S}_i$ with respect to the map frame ${\mathcal{M}}$. 
We empirically determined translational and rotational thresholds
which trigger a submap correction, respectively \SI{10}{\centi\meter} and
\SI{2.5}{\degree}. If the position/rotation change exceeds its threshold,
the submap is corrected:
\begin{equation}
||{^\mathcal{M}\hat{\mathbf{t}}_{\mathcal{S}_i}} - {^\mathcal{M}}\mathbf{t}_{\mathcal{S}_i}|| >
\lambda_\textup{update} 
\lor 
||{^\mathcal{M}\hat{\mathbf{R}}^{-1}_{\mathcal{S}_i}} {^\mathcal{M}\mathbf{R}_{\mathcal{S}_i}}|| >
\theta_\textup{update}
\end{equation}

Because we do not maintain a global map in our pipelines, this update
need only correct the root poses of the submaps, with no additional
global map fusion required.

\section{Experiments and Evaluation}
\label{sec:Experiments}

\secref{subsec:Runtime} presents the computation time and memory usage of the proposed system compared to OctoMap and Voxgraph.
In this section, we evaluate the global consistency of the proposed online 
elastic reconstruction pipeline using MultiresTSDF and
test the path planning application of MultiresOFusion.
To assess the level of global consistency achieved via submap elasticity, 
we compared the MultiresTSDF map with the ground truth of NCD.

For path planning, we tested on a dataset of a mobile
robot exploring an network of rooms in a underground mine, 
and used the MultireOFusion pipeline to create a high resolution
occupancy map so as to test suitability for path planning. 

\begin{figure}[t]
    \centering
    \includegraphics[width=\columnwidth]{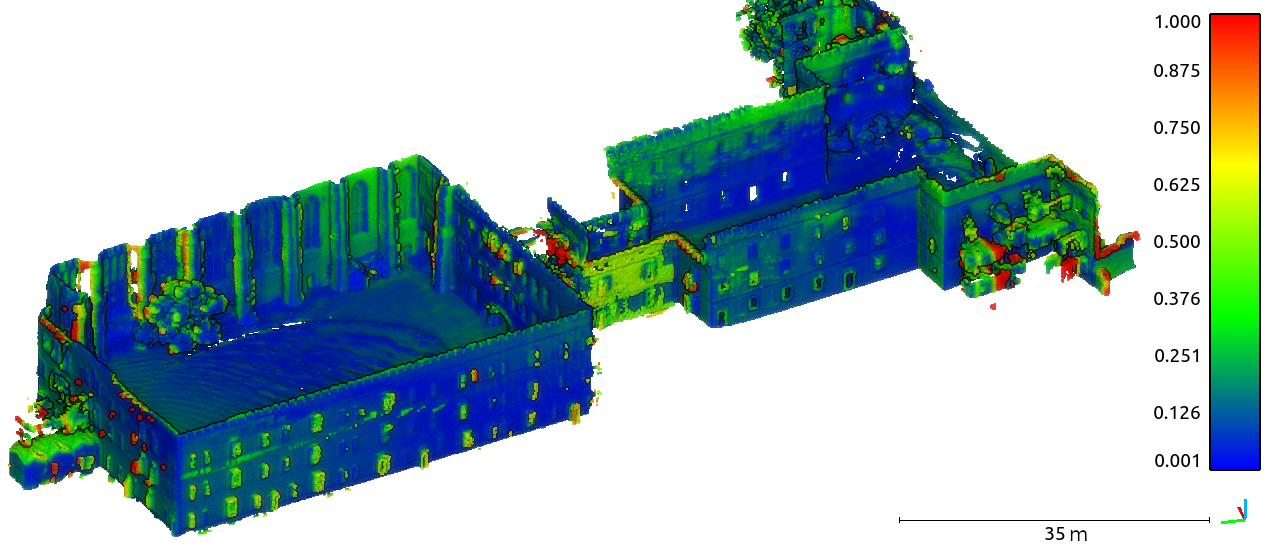}
    \includegraphics[width=0.8\columnwidth]{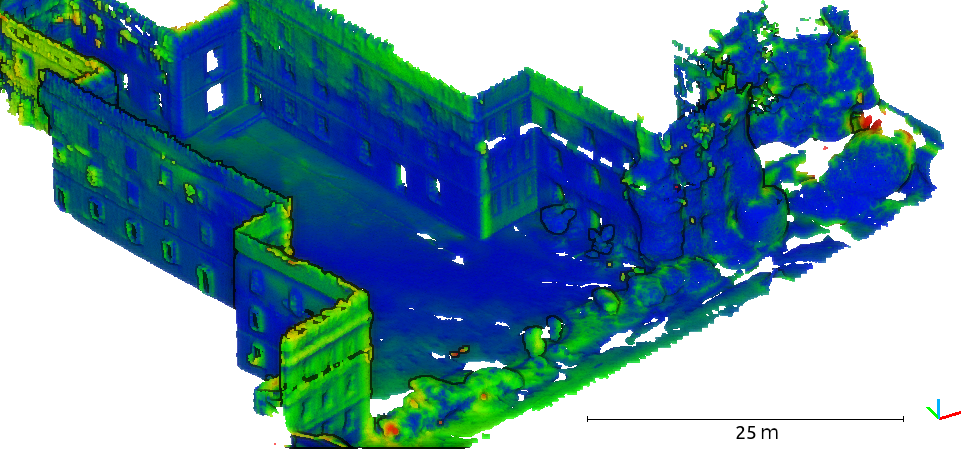}
    \caption{\small{Evaluation of reconstruction accuracy using point-to-point distance compared
            against
            ground truth.}}
    \label{fig:accuracy}
    \vspace{-6mm}
\end{figure}

\subsection{Reconstruction Accuracy}

In \figref{fig:accuracy}, we present the result of the MultiresTSDF
reconstruction on the NCD Long experiment~\cite{ramezani2020newer}. 
It is an exploration task
spanning nearly \SI{2.2}{\kilo\meter} in 
a large-scale
college campus.
Given the scale, the \SI{60}{\meter} maximum sensor range and \SI{6.5}{\centi\meter} voxel
resolution is needed for this reconstruction.

To evaluate the map quality, we fused all the submaps generated
at the end of the experiment into one global MultiresTSDF reconstruction and
extracted all the vertices of the mesh to form a full point cloud of
the explored environment. We then compared the extracted cloud with a ground
truth map collected with a Leica BLK 360, a survey grade tripod-based laser scanner.
The ground truth cloud was downsampled to \SI{5}{\centi\meter} to match
the resolution of our reconstruction.
We used CloudCompare\footnote{https://www.danielgm.net/cc/} to
align the ground truth and the reconstructed point cloud and 
to compute the point-to-point distance error between them. 

For about 90\% of the points, the distance error is less than \SI{50}{\centi\meter}. 
The tunnel and the doorway connecting each section of the campus 
were reconstructed with high accuracy. This is demonstrated in detail in
\figref{fig:ncd_short_tunnel}, where the finely detailed mesh of the tunnel
enables tasks such as robot path planning.
The reconstruction results of the ARCHE experiment are presented in the supplementary video. 

We note that the Ouster LiDAR used in the experiments is not as accurate as 
devices such as Velodyne, but Ouster's wide vertical FoV is valuable for 
our intended usecase of exploration and planning rather than mapping specifically.



\subsection{Path Planning in Underground Network}
\begin{figure}[t]
    \centering
    \includegraphics[width=0.9\columnwidth]{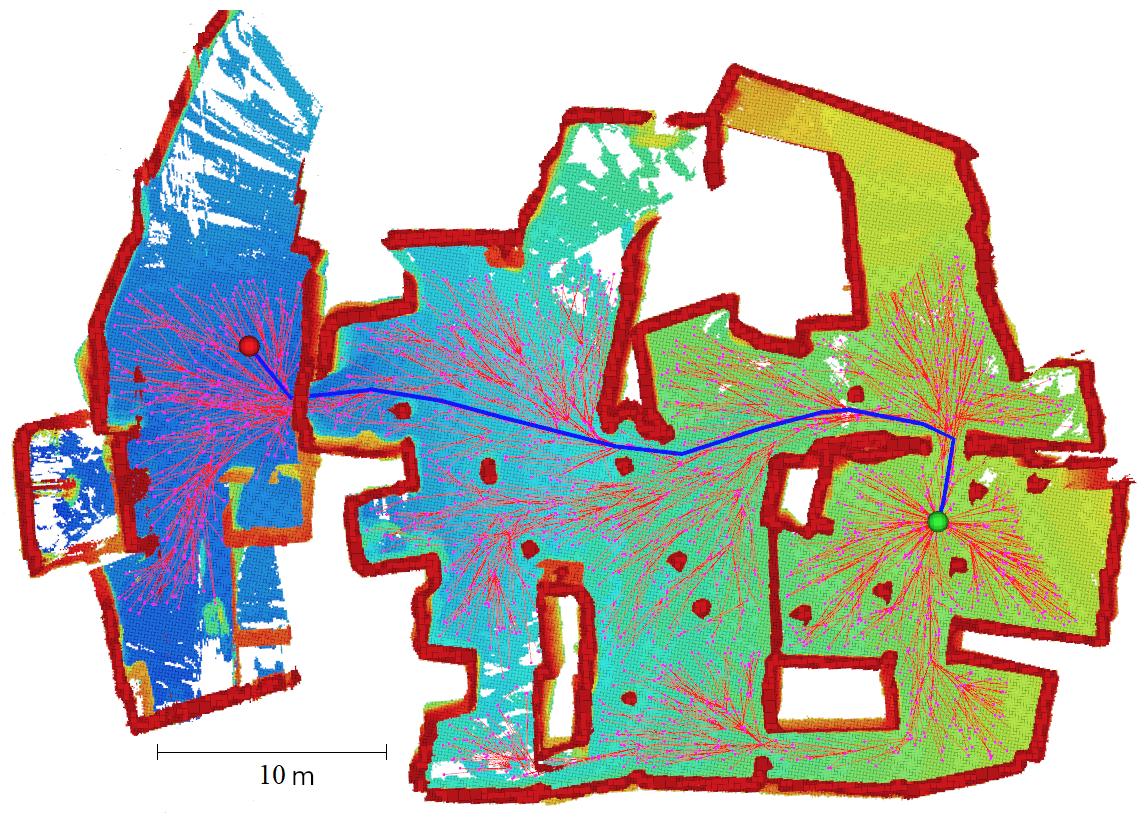}
    \caption{\small{Using the reconstruction result for path planning in an underground room
network. Green sphere - start; red sphere - end; 
red tree with magenta nodes - RRT*; 
blue trajectory - planned path.}}
    \label{fig:planning}
    \vspace{-6mm}
\end{figure}

To test the MultiresOFusion pipeline on a realistic
path planning application, we collected a dataset in an underground
mine consisting of a room network hewn from the rock. The
dataset was
collected with the same sensor platform as in NCD, but mounted on
a Husky wheeled robot.
We ran the pipeline with \SI{6.5}{\centi\meter} resolution
and \SI{60}{\meter} range again. 
The resultant occupancy map was then used by an
RRT*~\cite{Karaman2011} path planner to compute the shortest collision free path
between two locations. 
The result is presented in \figref{fig:planning}:
the volumetric reconstruction is highly detailed, giving clear definition
even in narrow doorways and corridors.
This allowed the path planner to find the optimal path to the goal despite
obstacles such a support pillars. 

\section{Conclusion and Future Work}
\label{sec:FutureWork}

To summarise, our proposed system can efficiently reconstruct 
large-scale environments at \SI{3}{\hertz} with high resolution 
(\SI{\sim5}{\centi\meter}) using long range LiDAR scans
(\SI{60}{\meter} max range). The core data structure of 
\textit{supereight} exploits sparsity of the environment with an adaptive resolution
representation
and allows for better scalability and efficiency in scan integration as
compared to
state-of-the-art mapping systems such as Voxgraph and OctoMap. 
Thanks to high resolution reconstruction around obstacles, 
our mapping result is suitable for navigation and path planning
even in challenging situations. 

We use submaps to introduce elasticity into the volumetric map. 
This allows the reconstruction to be corrected with SLAM loop closures 
during long exploration tasks. 
Our results demonstrate global consistency in the map
when compared to the ground truth. 
Graph clustering and submap fusion also further improve system's scalability.


To further improve the reconstruction accuracy, we plan to extend
\textit{supereight} to incorporate an undistorted spherical model to account
for LiDAR motion distortion and enable high speed operation. 
In addition, we aim to improve the submap fusion by applying 
a graph sparsification heuristic based on information theory, 
similar to \cite{Carlevaris2013, Mazuran2014NonlinearGS, Vallve2018},
so that the number of maps plateaus when a space is fully explored. 


\balance
\bibliographystyle{IEEEtran}
\bibliography{library}

\end{document}